\pdfoutput=1

\documentclass[11pt]{article}
\usepackage[]{natbib}
\usepackage[]{acl}
\usepackage{linguex}
\usepackage{times}
\usepackage{wrapfig}
\usepackage{graphicx}
\usepackage{stfloats}
\usepackage{latexsym}
\usepackage{amsmath}
\usepackage{hyperref}
\usepackage{caption}

\usepackage[T1]{fontenc}

\usepackage[utf8]{inputenc}

\usepackage{microtype}
\title{Plausibility Processing in Transformer Language Models:\\ Focusing on the Role of Attention Heads in GPT}

\author{Soo Hyun Ryu \\
    Department of Psychology \\ 
    University of Michigan \\
\texttt{soohyunr@umich.edu}}

\begin{document}
\maketitle
\begin{abstract}
The goal of this paper is to explore how Transformer language models process semantic knowledge, especially regarding the plausibility of noun-verb relations. First, I demonstrate GPT2 exhibits a higher degree of similarity with humans in plausibility processing compared to other Transformer language models. Next, I delve into how knowledge of plausibility is contained within attention heads of GPT2 and how these heads causally contribute to GPT2's plausibility processing ability. Through several experiments, it was found that: i) GPT2 has a number of attention heads that detect plausible noun-verb relationships; ii) these heads collectively contribute to the Transformer's ability to process plausibility, albeit to varying degrees; and iii) attention heads' individual performance in detecting plausibility does not necessarily correlate with how much they contribute to GPT2's plausibility processing ability. Codes are available at \href{https://github.com/soohyunryu/plausibility-processing-transformers}{github.com/soohyunryu/plausibility-processing-transformers}

\end{abstract}

\section{Introduction}
Transformers are attention-based neural network models  \cite{vaswani2017attention}, which have brought breakthroughs in the field of Natural Language Processing achieving state-of-the-art performance in diverse downstream tasks. Such great performance is thought to be attributed to Transformers' ability to build dependencies even between long-distant words which attention heads are developed for \cite{merkx2021human}. To be specific, unlike previous neural network language models (e.g., Simple Neural Networks or Recurrent Neural Networks) that have issues retaining linguistic information coming from distant tokens, attention heads in Transformers enable to represent the meaning of tokens by integrating their contextual information without losing information from distant tokens \cite{bahdanau2015neural}.
\par
Provided that Transformer language models consist of multiple attention heads that serve different roles, previous studies examined functions that individual attention heads serve and how language processing work is divided inside Transformers \cite{clark2019does, voita2019analyzing, vig2019multiscale,jo2020roles}. However, previous studies mostly focused on finding attention heads that process linguistic knowledge intrinsic to language systems such as morphosyntactic rules, and little attention has been paid to semantic knowledge, which requires much of world knowledge going beyond rules in language systems. 
\par
Consequently, we only have limited knowledge of how attention heads contribute to Transformers' general ability to process semantic knowledge. A number of studies \cite{bhatia2019distributed, bhatia2022transformer, ettinger2020bert, han2022human, misra2020exploring, misra2021language,  pedinotti2021did, peng2022copen, ralethe2022generic} examined how Transformers process semantic knowledge in comparison with humans, but their focus was mostly on the models' performance from the final hidden state without answering where the specific type of knowledge is preserved or processed in Transformer models. A few studies started investigating how world knowledge is stored in Transformers (e.g., \citet{meng2022locating} examined how GPT stores factual associations). However, the previous findings are yet generalizable to a broad range of semantic knowledge, and thus more studies are needed to understand how Transformers process other types of semantic knowledge. 
\par
In this regard, the present study aims to advance our knowledge of semantic knowledge processing in Transformer language models by closely investigating individual attention heads' ability in processing semantic plausibility and their causal contribution to Transformer's performance in plausibility processing. Among various types of plausibility, the especial focus of this paper is on the plausible relation between nouns and verbs. While recognizing the importance of considering a broader array of semantic knowledge in future studies, I made this specific choice because the objectives of the present paper are to demonstrate  a set of attention heads can be specialized for specific type of semantic knowledge and to introduce a set of analyses that can be used to probe attention heads' role in processing semantic knowledge.
\par The semantic plausibility of the relationship between nouns and verbs can be determined by the degree to which semantic features of nouns and verbs match, as shown in sentences in (1) from \citet{cunnings2018retrieval}. For instance, in (1a), the syntactic dependent (\textit{plate}) of the verb (\textit{shattered}) has a feature [+shatterable], which builds a plausible relation with the verb (\textit{shattered}). In (1b), however, the syntactic dependent \textit{letter} does not have a feature [+shatterable], and thus it is semantically implausible dependent of the verb (\textit{shattered}).

\ex.
\a. {Sue remembered the \textbf{plate} that the butler \textbf{shattered} ...}
\b. {Sue remembered the \textbf{letter} that accidentally \textbf{shattered} ...}

\par
In order to examine how such knowledge is preserved and processed inside Transformer-based language models, this paper answers the following questions: (i) How similar are Transformer's plausibility processing patterns to humans'?; (ii) How sensitive is each of the attention heads in  Transformers to plausibility relation?; and (iii) How do these heads make causal effects on Transformers' ability to process semantic plausibility?
\par
\par 

After comparing patterns in plausibility processing between a group of Transformer-based language models and humans, it was found that GPT2 tends to process the plausibility between nouns and verbs in a way that is more similar to humans than other language model types. Several follow-up experiments that especially focus on GPT2 answered the last two questions. Specifically, it was uncovered that GPT2 has a set of attention heads that detect semantic plausibility, which are relatively diffusely distributed from the bottom layers to the top layers and that they exert causal effects on Transformers' semantic plausibility processing ability. GPT2's plausibility processing ability almost disappeared when the plausibility-processing attention heads are pruned, but the effects of removing a plausibility-processing attention head was not balanced nor proportional to the attention heads' performance in detecting plausible nouns. Rather, it was found that a single attention head accounts for most of plausibility processing ability of GPT2. 
\par
In what follows, I will provide a background that relates to the questions I address in this paper. In Section \ref{sec: human-gpt}, I will compare Transformer-based language models' and humans' sensitivity to the plausibility of the relation between nouns and verbs. In Section \ref{sec:plausibility-processor}, I will conduct an experiment to find attention heads that can detect semantic plausibility knowledge and examine how they are distributed inside the model. In Section \ref{sec:causal-effect},  it will be examined how individual attention heads collectively make causal effects of  on Transformers' sensitivity to plausibility. In Section \ref{sec:conclusion}, I will summarize the results and discuss the limitations of the study. 
\section{Background}
\paragraph{What roles do attention heads serve?}
There have been a lot of studies that attempted to explain the language processing mechanism in Transformers with analyzing functions that distinct attention heads serve \cite{voita2019analyzing,vig2019multiscale,clark2019does,jo2020roles}.
Specifically, \citet{voita2019analyzing} found attention heads specialized for a position, syntactic relation, rare words detection; \citet{vig2019multiscale} found attention heads specialized in part-of-speech and syntactic dependency; \citet{clark2019does} found attention heads specialized in coreference resolution; and \citet{jo2020roles} examined how linguistic properties at the sentence level (e.g., length of sentence, depth of syntactic trees and etc.) are processed in attention heads. 
\par
Despite numerous attempts in examining the roles of attention heads, the focus has been mostly on linguistic knowledge intrinsic to language systems which does not require much world knowledge that is indispensable for 
semantic knowledge processing. Thus, it needs to be closely examined how Transformers preserve and process such knowledge that facilitates sentence processing. 
\par
\paragraph{How do we learn attention heads are specialized for certain linguistic knowledge?}
In previous studies, attention heads are considered to be able to process a certain type of linguistic knowledge if attention distribution patterns in the attention heads are consistent with the linguistic knowledge \cite{voita2019analyzing, vig2019analyzing, ryu2021accounting}. However, such regional analysis does not explain how much contribution attention heads make to Transformers' ability to process linguistic knowledge because such information from the attention heads may fade away or be lumped along with the information flows - from bottom layers to top layers - eventually making little contribution to Transformers' ability to process the linguistic knowledge. Thus, to rigorously confirm the role of attention heads in processing a certain type of knowledge, it is crucial to analyze the causal effects that they make on Transformer's ability to process linguistic information \cite{belinkov2019analysis,meng2022locating,vig2020causal}.
\par
In this sense, this paper will not only examine which attention heads can form attention distributions that are consistent with semantic plausibility knowledge, but also examine how much influence the attention heads can exert on Transformers' general ability to process plausibility.

\section{Comparison between humans and Transformer language models in plausibility processing patterns}
\label{sec: human-gpt}
This section examines how a set of Transformer language models process plausibility of noun-verb relations in comparison with human data. 
\subsection{Data}
In \citet{cunnings2018retrieval}, it was investigated how the degree of noun-verb plausibility affects the way humans process sentences. There are 32 sets of sentences with varying not only the plausibility of dependent-verb relations but also the plausibility distractor-verb relations\footnote{In experiments with language models, I removed sets of sentences whose tokens of interest are not recognized as a single token by the tokenizer.}.

\ex. 

\a. \emph{plausible - plausible} \\
... that the \textbf{plate} that the butler with the \underline{cup} accidentally \textbf{shattered} ...\vspace{0.15cm}
\b. \emph{plausible - implausible} \newline ... that the \textbf{plate} that the butler with the \underline{tie} accidentally \textbf{shattered} ...\vspace{0.15cm}
\c. \emph{implausible - plausible} \newline ... that the \textbf{letter} that the butler with the \underline{cup} accidentally \textbf{shattered} ...\vspace{0.15cm}
\d. \emph{implausible - implausible} \newline ... that the \textbf{letter} that the butler with the \underline{tie} accidentally \textbf{shattered} ...

\subsection{Method}
\citet{cunnings2018retrieval} measured the degree of difficulty that people have when processing a certain noun-verb pair with reading times that are measured at verb\footnote{The original paper also talks about the spillover region following the verbs of interest, but this study focuses on  the reading times (total viewing times) measured at the verb region.} (\textit{shattered} in (2)). To compare humans' responses with Transformer language models, I computed surprisals \cite{hale2001probabilistic, levy2008expectation}, also measured at verbs, as a metric that represents processing difficulty of the model, given a large set of evidence manifesting that surprisals computed from neural network language models can simulate human sentence processing patterns \cite{futrell2019neural, michaelov2020well,van2021single, wilcox2020predictive}. 
\par
Surprisal is a term that estimates the degree of the unexpectedness of tokens given their preceding context, which is computed by taking the negative log probability of a token conditioned on its preceding words (See Equation (\ref{eq:surprisal})). In neural network language models, the surprisal of a word is computed using the softmax-activated hidden state before consuming the word \cite{wilcox2018rnn}.
\begin{equation}\tag{A}
    {Surprisal}(w) = -{\log_2}P(w|h)
\label{eq:surprisal}
\end{equation}
where \textit{h} is the softmax-activated hidden state of the sentence before encountering the current word. 
\par
Both reading times and surprisals measured at verbs are expected to be greater in sentences with implausible nouns than in ones with plausible nouns since it is less likely to anticipate a certain verb after encountering a noun in an implausible relationship with the verb. 
\par
A set of Transformer language models to be tested includes ALBERT \cite{lan2019albert}, RoBERTa \cite{liu2019roberta}, BERT \cite{kenton2019bert}, and GPT2 \cite{radford2019language}. The versions of models that are tested have 144 attention heads, which are spread across 12 layers with 12 attention heads each. Models are accessed through Huggingface \cite{wolf2019huggingface}.
\subsection{Results}

\begin{figure*}[h!]
\centering
\includegraphics[scale=0.34]{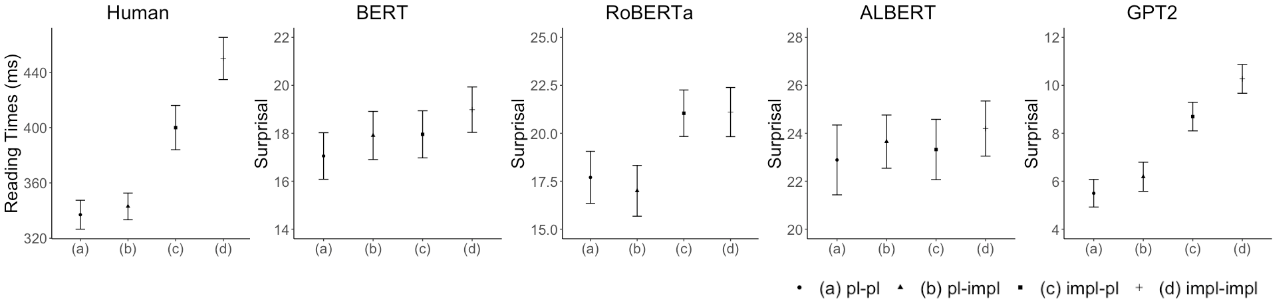}
\caption{Surprisals computed from Transformer language models and reaction times from human subjects for processing different types of noun-verb pairs. Human reading times are from \citet{cunnings2018retrieval}. Shapes at the center and intervals for each condition represent means and standard errors.}
\label{fig:human-gpt-comparison}
\end{figure*}
As shown in Figure \ref{fig:human-gpt-comparison}, GPT2 exhibits the highest level of similarity to humans in processing the plausibility of noun-verb pairs, in comparison to other Transformer-based language models.
\par
In addition, further statistical analysis using regression models supports GPT2's similarity with humans in plausibility processing. First, significantly lower processing difficulties are observed when syntactic dependents are in a plausible relationship with the verb than when they are in an implausible relation for both human (estimate = .11 , SE = .01, \textit{t} = 9.26, \textit{p} < .001) and GPT2 (estimate = 4.81 , SE = .84, \textit{t} = 4.86, \textit{p} <.001).
\par
Also, GPT2 showed marginally significant plausibility effects even with distractors that do not form a dependency relation with the verb (estimate = 1.57, SE = .84, \textit{t} = 1.87, \textit{p} = .06) (i.e., processing difficulties are greater in (b) and (d) than in (a) and (c)), similar to the human data where significant plausibility effects from distractors are found (estimate = .04, SE = .13, \textit{t} = 2.85, \textit{p} < .05)\footnote{Plausibility effects observed for distractors in GPT2 and humans are due to the illusion of plausibility \cite{cunnings2018retrieval}: even distractors that cannot build syntactic dependency with cues (verbs) can be illusorily considered as the syntactic dependents, causing moderate plausibility effects while sentence processing.}. 
\par
Being inconsistent with the human reading time data that show the interaction effects of dependent-plausibility and distractor-plausibility (estimate = .02, SE = .01, \textit{t} = 2.29, \textit{p} < .05), GPT2 data do not show significant interaction effects (estimate = .89, SE = 1.19, \textit{t} = .75,   \textit{p} = .46). This absence of evidence for interaction effects in GPT2  may be due to the difference in sample sizes, which can impact the level of statistical significance.  It would be possible to observe the interaction effects with the increased data size especially given a trend of interaction in GPT2: the surprisal difference between (a) and (b) is smaller than the surprisal difference between (c) and (d), consistent with human data. For the statistical results from other Transformer-based language models, see Appendix \ref{sec:appendixAA}.
\subsection{Discussion}
Compared to other language models, GPT2 is found to process plausibility between nouns and verbs in a similar way as humans do. While more rigorous study is required to explain the origin of GPT2's supeiror performance in simulating human plausibility processing patterns, I assume that the GPT2's similarity to humans arises from the psychological plausibility of its decoder-only architecture. In particular, it processes sentences incrementally much like the way humans process sentences (i.e., it constructs the meaning of a certain word only given its prefix, without any influence from the `unseen' next coming words), unlike other types of language models that are tested exploit bidirectional processing (i.e., it process each word of sentences not incrementally, but integrating both preceding and following words.) 

\par
Given that GPT2 shows the most similar patterns as humans in processing plausibility of noun-verb relations, the following sections will examine the role that attention heads in plausibility processing, focusing on the GPT2 model. 

\section{Plausibility processing attention heads in GPT2}
\label{sec:plausibility-processor}
 \par
This section will examine whether GPT2 has a specific set of attention heads that can sensitively detect plausiblity of noun-verb relations, irrespective of syntactic dependency relation. Experimental stimuli were the same as previous experiment. 
\par

\subsection{Method}
In GPT2's attention heads, each token allocates different amounts of attention to previous tokens depending on the relevance of the two tokens\footnote{The relevance can be defined in terms of functions that attention heads serve. For instance, if an attention head is specialized for detecting \textit{subject-verb} dependency relation, the amount of attention can reflect how likely two tokens are in the \textit{subject-verb} relationship \cite{voita2019analyzing}}.  
\par
With such a property of Transformers, the capacity of attention heads in detecting plausibility is measured in terms of \textit{accuracy} that indicates how likely the plausible noun is to get higher attention than the implausible noun in a certain attention head (See Equation (\ref{eq:accuracy})). 
\begin{equation}\tag{B}
\begin{aligned}
{Accuracy_{lh}}= \hspace{4.5cm}  \\
\hspace{-1cm}\frac{\sum_{j=1}^{k}[{Attn(pl_{j}, v_{j}) > Attn(impl_{j}, v_{j})]}}{{k}}
\end{aligned}
\label{eq:accuracy}
\end{equation}
, where \textit{lh} refers to the location of attention heads (\textit{h} for the \textit{h}th head in the \textit{l}th layer), \textit{j} refers to the sentence id, $pl_{j}$ and $impl_{j}$ refer to the plausible and implausible nouns to be compared in the \textit{j}th sentence set, $v_{j}$ refers to the verb in the \textit{j}th sentence, and \textit{k} is the number of sentence sets.



\par In order to ensure that the heads do not particularly work for tokens that form syntactic dependency but work for semantically related tokens, I measured the accuracy not only using pairs of syntactic dependents (\textit{plate} vs. \textit{letter} in (2)), but using pairs of distractors (\textit{cup} vs. \textit{tie} in (2)). Considering both of noun types enabled to find attention heads that can judge the plausibility between nouns and verbs regardless of syntactic compatibility between them. Thus, there are four comparisons between \textit{plausible} and \textit{implausible} conditions for each set of sentences: (pl-pl vs. pl-impl), (impl-pl vs. impl-impl), (pl-pl vs. impl-pl), (pl-impl vs. impl-impl), where the first and the second corresponds to syntactic dependents and distractors, respectively.

\subsection{Results}          

I consider attention heads are able to process plausible relationships between nouns and verbs when their accuracy in identifying appropriate nouns surpasses the chance level, having the cutoff as 70\% at my discretion. To select attention heads that can process the semantic plausibility regardless of the syntactic dependency relation between the noun and the verb, I consider attention heads whose accuracies are greater than 70\%  in both noun types.
\par
With such criteria, eighteen attention heads are recognized to be able to process plausibility:  [(0, 1), (0, 5), (0, 10), (1, 5), (1, 6), (1, 11), (3, 0), (4, 3), (4, 4), (4, 10), (5, 10), (5, 11), (6, 6), (7, 1), (7, 9), (8, 3), (8, 10), (9, 4), (10, 7)], where the first numbers refer to indexes of layers and the second refer to indexes of heads (i.e., (\textit{i}, \textit{j}) refers to the \textit{j}th head in the \textit{i}th layer.)) Among the attention heads that are found to process semantic plausibility, two attention heads - (1, 6) and (5, 10) - especially show noteworthy performance in detecting plausible, achieving 95\% of accuracy. Please refer to Appendix \ref{sec:appendixA} to see the values from each head. 
 \par

\subsection{Discussion}
This section showed that a set of attention heads are particularly good at processing semantic plausibility between nouns and verbs. Such plausibility processing ability seems independent of their ability to process syntactic dependencies since their ability to process plausibility is not limited to processing syntactic dependents of verbs, but it is also applicable to distractors that do not form any syntactic dependencies with verbs.
\par
Unlike attention heads specialized for processing a certain syntactic relation and superficial linguistic information such as word position or word rarity is clustered in a relatively small region \cite{voita2019analyzing}, it seems that the components that process semantic plausibility are relatively evenly distributed across twelve layers and take up an even greater region: 18 attention heads out of 144 attention heads in the GPT2-small model. In the next section, it will be discussed how these plausibility-processing attention heads collectively exert causal effects on GPT2's plausibility-processing ability. 

\section{Causal effects of plausibility-processing attention heads on GPT2's plausibility sensitivity}
\label{sec:causal-effect}
In the previous experiment, attention heads capable of detecting plausible relations between nouns and verbs were found. The present section examines how such attention heads make causal influence on GPT2's sensitivity to plausibility between nouns and verbs. In particular, I attempt to answer two questions: (i) How GPT2's responses to plausible/implausible verb-noun pairs change when plausibility-processing attention heads are removed? and (ii) How does GPT2's plausibility-sensitivity change as attention heads are gradually pruned?
\subsection{Influence of a set of plausibility-processing heads to plausibility sensitivity}
\label{sec:sensitivity}
In this study, I examine how GPT2's responses to plausible and implausible noun-verb relations change when the plausibility-processing heads are removed.


\subsubsection{Method} Surprisals are computed from two models: i) GPT2 without plausibility-processing heads and ii) GPT2 after removing the same number of attention heads as i), but the heads to prune selected randomly. I included the random-removal model to see whether the disappearance of the plausibility sensitivity in GPT2 is simply attributed to taking away some portion of the information in GPT2, or it is caused by specifically removing plausibility processors. In order for reliability, we used 100 different random attention head sets for ii), and computed the average of surprisals from the 100 models.
\par
Attention heads were pruned by replacing attention values with zeros, following \citet{michel2019sixteen}.
\subsubsection{Results}
\begin{figure}[h]
    \includegraphics[scale=0.25]{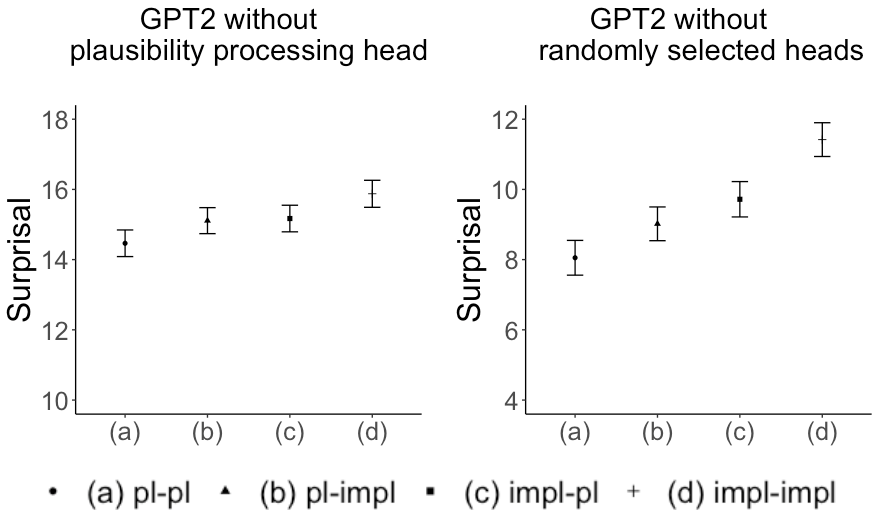}
\caption{Surprisals computed from GPT2s after removing different sets of attention heads and reaction times from human subjects for processing different types of noun-verb pairs.}
\label{fig:ex2-results}
\end{figure}

\par When removing the plausibility processing attention heads (left in Figure \ref{fig:ex2-results}), no plausibility effects are found for syntactic dependents (estimate = .77, SE = .53, \textit{t} = 1.43, \textit{p} = .15) and for distractors (estimate = .71, SE = .54, \textit{t} = 1.32, \textit{p} = .19). Also, no interaction effects are found (estimate = 0.06,  SE = 0.76, \textit{t} = 0.08, \textit{p} = 0.94)
\par 
Importantly, such a decrease is not the effect that is caused by simply removing some random components in GPT2. When randomly selected eight-teen attention heads are pruned (right in Figure \ref{fig:ex2-results}), the GPT2 model better simulates human responses in processing plausibility. In this case, the significant plausibility effects are observed both in syntactic dependents (estimate = 2.40, SE = .69, \textit{t} = 3.46, \textit{p} < .001 ) and in distractors (estimate = 1.70, SE = .69, \textit{t} = 2.45, \textit{p} < .05), although interaction effects are not found as well (estimate = 0.73, SE = 0.98, \textit{t} = 0.75 \textit{p} = 0.46).
\par

\subsection{Gradual changes in GPT2's plausibility sensitivity as attention heads are pruned}
\label{subsec:gradual}
The previous section examined how the set of plausibility-processing attention heads influences GPT2's responses to plausible or implausible noun-verb relations. Though it was shown that plausibility processing attention heads collectively contribute to GPT2's ability to process plausibility unlike other sets of attention heads, it is unanswered how individual attention heads contribute to GPT2's plausibility-processing ability. Do they have balanced contributions to GPT2's ability to process plausibility? Or, is it that only a small set of plausibility-processing attention heads account for most of the plausibility-processing ability of GPT2?  In order to answer these questions, the following experiment investigates how GPT2's general sensitivity to plausibility gradually changes as attention heads are  pruned one by one. 
\subsubsection{Method}
This study operationalizes GPT2's plausibility sensitivity as the difference in \textit{surprisals}  measured at the verbs of interest (`\textit{shattered}' in (2)) in sentences with plausible nouns and in ones with implausible nouns as shown in Equation (\ref{eq:pl}).
\begin{equation}\tag{C}
\begin{aligned}
    {Plausibility\ Sensitivity} = \hspace{2cm} \\
    {surprisal_{impl}}(verb) - {surprisal_{pl}}(verb)
\end{aligned}
\label{eq:pl}
\end{equation}
, where $\mathrm{surprisal_{pl}}$(verb) and  $\mathrm{surprisal_{impl}}$(verb) refer to surprisals measured at the verb in a sentence with a plausible noun and in a sentence with an implausible noun, respectively. 
\par I computed two plausibility sensitivities: one that compares surprisals at verbs when having plausible syntactic dependents of verbs in sentences and having implausible syntactic dependents (\{(c)+(d)\} - \{(a)+(b)\}) and the other that compares surprisals when having plausible distractors of verbs and implausible distractors (\{(b)+(d)\} - \{(a)+(c)\}). 
\par
Both types of plausibility sensitivities are measured at each point after gradually removing a plausibility processing attention head one by one. Attention heads were pruned in decreasing order of their accuracies\footnote{I used the average values of accuracies for dependents and for distractors that were computed in Section 3.} in detecting plausible nouns over implausible nouns.

\subsubsection{Results}
\begin{figure*}[t!]
    \centering\includegraphics[scale=0.35]{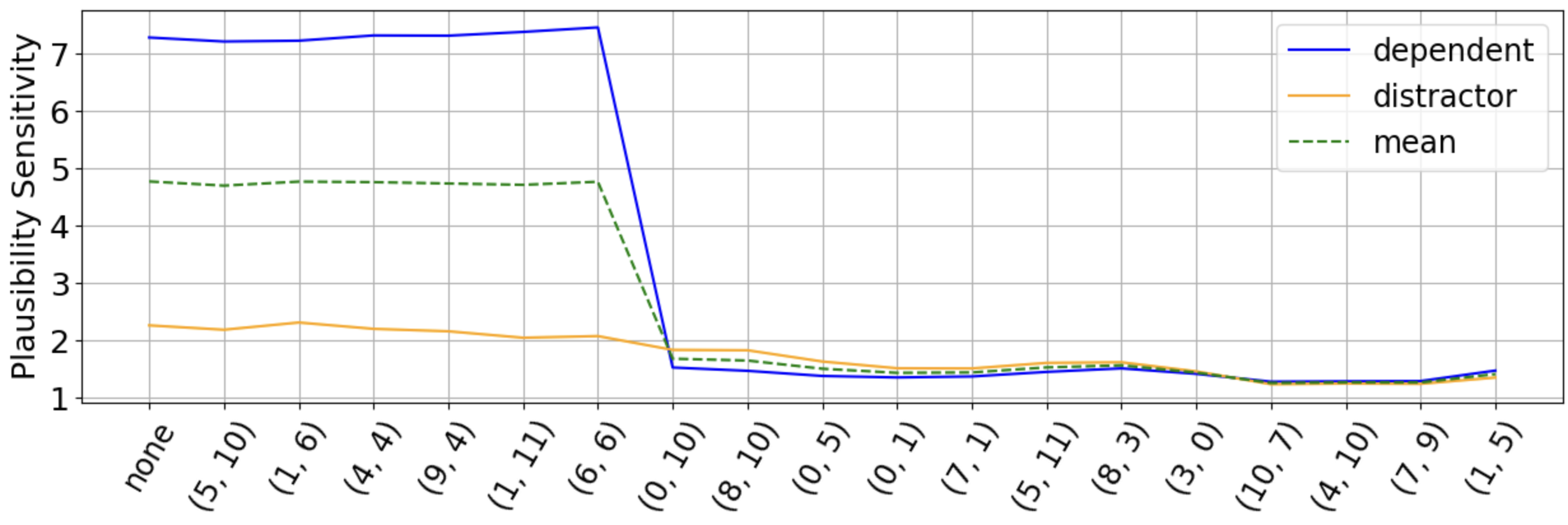}
\caption{Changes in plausibility sensitivity by noun types as attention heads are gradually pruned. X-axis indicates plausibility-processing attention heads that are pruned at a certain point.}
\label{fig:ex3-results}
\end{figure*}

Figure \ref{fig:ex3-results} plots how the plausibility sensitivities for both types of noun-verb relations change as plausibility-processing attention heads are removed gradually.
\par When it comes to the plausibility sensitivity for distractors, the changes seem to be continuous. Such patterns suggest that the set of plausibility processing attention heads make a collective contribution to plausibility effects for distractors. Such collective contribution that plausibility processing attention heads make is especially supported by the fact that the gradual decrease in plausibility sensitivity over the course of removing 18 attention heads eventually led to the elimination of the statistically significant plausibility effects for distractors as observed in Section \ref{sec:sensitivity}.
\par
\begin{figure}[h]
\centering\includegraphics[scale = 0.35]{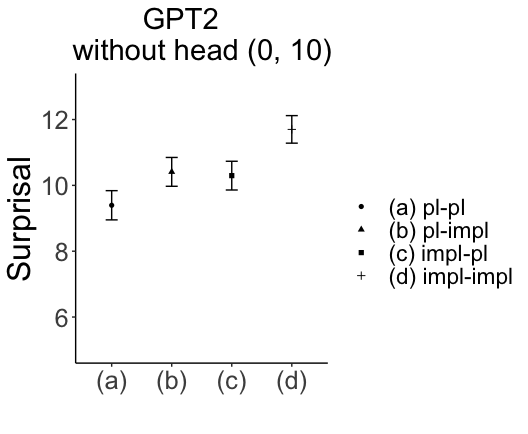}
\caption{Surprisals by conditions computed with the GPT2 without a single attention head (0, 10)}
\label{fig:ex3-app}
\end{figure}

In contrast, the sensitivity to plausibility for the relation between syntactic dependents and verbs shows a drastic decrease upon the removal of the attention head (0, 10). The effect from the removal of the head (0, 10) shows that this particular head exerts a huge amount of causal effects on GPT2's general sensitivity to plausible relations between syntactic subjects and verbs\footnote{The drastic drop after the removal of the head (0, 10) was also found when attention heads are removed in random order.}. Figure \ref{fig:ex3-app} confirms that the head (0, 10) causes a huge amount of causal contribution on GPT2's plausibility processing ability since it reduces the difference in surprisals between plausible conditions and implausible conditions, though it does not alone eliminate the significance in plausible effects for syntactic dependents (estimate = 1.29, SE = 0.61, \textit{t} = 2.10, \textit{p} < .05) or for distractors (estimate = 1.40, SE = 0.61, \textit{t} = 2.29, \textit{p} < .05).
\par
One additional interesting finding is that the general level of surprisals upon the removal of the attention head (0, 10) increases considerably regardless of the condition. For instance, the removal of the single attention head (0, 10) increases surprisals by 2.79 bits on average across the four conditions, which seems to be huge given that the randomly selected 18 attention heads only led to the 1.89 bits of increase. Such trends indicate one possible explanation of the role of the head (0, 10): it contributes to GPT2's general ability to predict the next word, and such impact arises in any sentence, not only in the sentences that require plausibility-processing. In the next section, further analysis on the role of the attention head (0, 10) will be provided to address such a possibility.

\subsection{Further analysis on the role of the attention head (0, 10)}
To better understand the origin of GPT2's plausibility processing ability, the present study aims to further examine the role of (0, 10) that make great contribution to plausibility sensitivity in GPT2 . In particular, I examine whether the (0, 10) is only specialized for semantic plausibility or is responsible for predicting next words in general sentences which leads to influence plausibility processing.
\subsubsection{Method}
Perplexity in Equation (\ref{eq:perplexity}) is the average value of surprisals computed from every tokens in corpus, which can be used to estimate the predictive power of language models in predicting next words given preceding context \cite{goodkind2018predictive}. 
\begin{equation}\tag{D} 
\begin{aligned}
    {Perplexity(LM)} = 
    {\frac{1}{M}\sum_{i=1}^{m}log_{2}P(w_i|h)}
\end{aligned}
\label{eq:perplexity}
\end{equation}
, where \textit{i} is the index of words, \textit{m} is the number of words in corpus, and \textit{h} refers to the softmax-activated hidden state of the preceding context.
\par
To examine how the general predictive power gets affected by the removal of the head (0, 10) in comparison with the removal of other heads, I computed the perplexities of GPT2 after removing each of 144 attention heads and compared those values. \citet{anderson}'s ``The Money Box'' story which has 41 sentences was used to compute perplexities. 
\subsubsection{Results}
\begin{figure}[h]
\includegraphics[scale = 0.35]{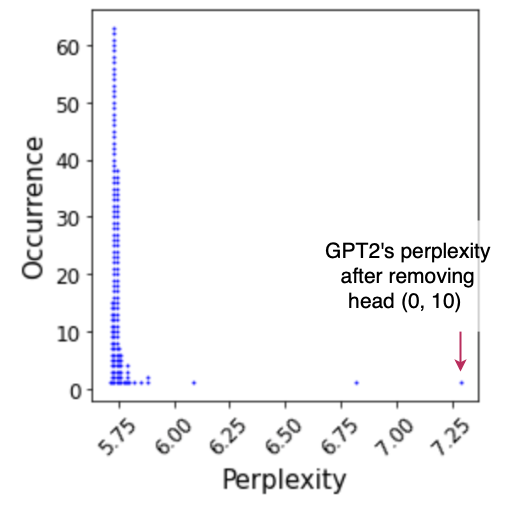}
\caption{Histogram of 144 perplexities of GPT2, each of which is computed after removing single attention head}
\label{fig:pp-histo}
\end{figure}
The perplexity of GPT2 with the entire set of attention heads was 5.47. In most of the cases, the removal of a single head does not seem to considerably affect GPT2's perplexity, since the perplexity remains to be in a similar range after the removal as shown in Figure \ref{fig:pp-histo}\footnote{For 95\% of attention heads, the perplexities change by less than 0.1 bit after the removal.}. However, it is clear that the removal of the head (0, 10) seriously harms the general predictive power of GPT2 because the perplexity becomes 7.27 after removing it, which is much greater compared to the most of other attention heads. This suggests that the head having the greatest influence on GPT2's plausibility processing ability is not specifically specialized for plausibility processing, but rather the attention head contributes to the general predictive power of any kind of sentence. 

\subsection{Discussion}
Results of this section suggest plausibility processing in GPT2 requires a collective contribution from a large set of plausibility processing attention heads, given that plausibility sensitivity decreases continuously as attention heads are gradually pruned.
\par
At the same time, however, it was also shown that the amount of causal effects that each attention head makes are highly imbalanced since the attention head (0, 10), which contributes to GPT2's general predictive power, leads to a significantly more drastic decrease in plausibility sensitivity for dependents than other heads. Taken together, although a single attention head can account for a great portion of the plausibility effects, other plausibility-processing attention heads make an additional contribution to GPT2's plausibility-processing ability. 
\par
Interestingly, the head (0, 10) did not achieve noteworthy performance in detecting plausible nouns over implausible nouns in Section \ref{sec:plausibility-processor}. This suggests that analyzing the causal effects each attention head makes is essential to understanding the role that attention heads serve, provided that the performance that each attention head shows in processing particular linguistic information does not necessarily align with how much it contributes to the model's performance in processing the specific information. 
\par
In addition, how the plausibility-processing attention heads affect Transformers' general ability needs to be investigated in relation to other attention heads that are specialized for different linguistic knowledge. This is especially the case given the findings that the way plausibility sensitivity decreases along with the gradual heads-pruning varies by the relation types that nouns build with verbs (i.e., syntactic dependents or distractors), which must be handled by different attention heads. 


\section{Conclusion \& Limitations}
\label{sec:conclusion}
The present study has shown how semantic plausibility is processed in Transformer language models, especially focusing on the role of attention heads. First, I demonstrated that GPT2, whose decoder-only architecture is more aligned with the way humans process sentences, shows greater similarity to humans in plausibility processing compared to other Transformer-based models such as BERT, RoBERTa and ALBERT.  Then, a series of experiments showed a set of attention heads are found to process plausibility, and those heads are diffusely distributed across 12 layers in GPT2. Moreover, it was observed that they make imbalanced but collective causal contributions to GPT2' plausibility-processing ability, which establishes the importance of causal effect analysis in attention-head-probing studies.
\par Although the results provide a window into how Transformers process semantic knowledge of plausibility, this study has a few limitations to be addressed in future studies. First, the scope of the study is restricted to the plausibility of noun-verb relations although there exist many different types of semantic knowledge. This limitation stems from the present paper's intention to `initiate' an exploration of Transformers' attention heads in handling of semantic knowledge and to exploit diverse and robust techniques for the exploration, rather than serving as a definitive endpoint that accounts for an exhaustive set of semantic knowledge.  However, future investigations should expand the current study’s scope for better generalizability.
\par
Moreover, the study does not detail how attention heads interact with other components like hidden states across layers or feed-forward perceptrons. Such details would be essential in enhancing our understanding of the attention head roles in plausibility processing by elucidating how these heads impact Transformer models' plausibility processing ability. As such, subsequent studies should delve deeper into these interactions for a more accurate understanding of their role in semantic knowledge processing.
\par
As these limitations are addressed, I anticipate further advancements in explaining Transformer models' capacity for semantic knowledge processing, founded on the novel findings and methods introduced in this study.
\section*{Acknowledgements}
This research took place as part of EECS 595 Natural Language Processing, a course taught by Joyce Chai at the University of Michigan in the fall term of 2022. I am truly grateful for the invaluable insights shared by all of my class instructors. Additionally, I extend my gratitude to the members of the Computational Cognitive Science Lab at the University of Michigan - Richard Lewis, Logan Walls, Yuxin Liu, Andrew McInnerney, Sean Anderson and Sarah Liberatore - for their instructive suggestions and guidance. I am also deeply appreciative of the four anonymous reviewers at the ACL Rolling Review for their productive feedback, which significantly enhanced the quality of the paper.

\bibliography{anthology,custom}
\clearpage
\appendix

\section{Statistical analysis on plausibility effects}\label{sec:appendixAA}
\begin{table*}[!t]
\centering
\caption{Statistical analysis on plausibility effects in human and Transformer-based language models.}
\label{tab:example2}
\begin{tabular}{|c|c|c|c|c|c|c|}
\hline
\multicolumn{2}{|c|}{}& Human  & BERT & RoBERTa & ALBERT & GPT2\\
\hline
\multicolumn{2}{|c|}{Difficulty measurement}& reading times & \multicolumn{4}{|c|}{ surprisals}\\
\hline
\hline
& estimate &  .11 & 1.10 & 4.11 & .55&  4.81\\
\cline{2-7}
Plausibility effects  & SE &  .01 &1.38 & 1.83 &1.77 &.84\\
\cline{2-7}
 (syntactic dependents) & \textit{t} & 9.26 & .78 & 2.24 & .31 & 4.86 \\
\cline{2-7}
 & \textit{p} &\textbf{ <.001} & .44 & \textbf{<.05} & .76 &\textbf{ <.001} \\
\hline

& estimate &  .04 & 1.03 & .06 &.87&  1.57\\
\cline{2-7}
Plausibility effects  & SE &  .13 & 1.38& 1.83 &1.77& .84\\
\cline{2-7}
 (distractors) & \textit{t} & 2.85 & .75 & .03 &.49& 1.87 \\
\cline{2-7}
 & \textit{p} & \textbf{<.05 }& .46 & .97 &.62& \textbf{< .10} \\
\hline
& estimate &  .02 & .17 & .76 &.11&  .89\\
\cline{2-7}
Interaction effects  & SE &  .01 &1.95 & 2.59 &2.50&1.19\\
\cline{2-7}
 ( dependents $\times$ distractors) & \textit{t} & 2.29 &.09& .29 &.04& .75 \\
\cline{2-7}
 & \textit{p} & \textbf{<.05}  & .93 & .77 &.96& .46 \\
\hline
\end{tabular}
\end{table*}
In order for quantitative analysis on how well Transformer language models simulate plausibility effects found in human data \cite{cunnings2018retrieval}, linear regression models for language model data were fit with the following equation: $ surprisal \sim \newline subject\_plausibility*distractor\_plausibility$. 
\par
The results are shown in Table \ref{tab:example2}. Results for human data are from \citet{cunnings2018retrieval}.

\section{Scores for detecting the plausible noun-verb relations by attention heads}\label{sec:appendixA}
The performance of attention heads in selecting the plausible nouns in relation with verbs over the implausible ones was measured in terms of \textit{accuracy} in the main text. The details of the method are provided in Section \ref{sec:plausibility-processor}.
\par
\begin{figure*}[!h]
\centering
\includegraphics[scale = 0.28]{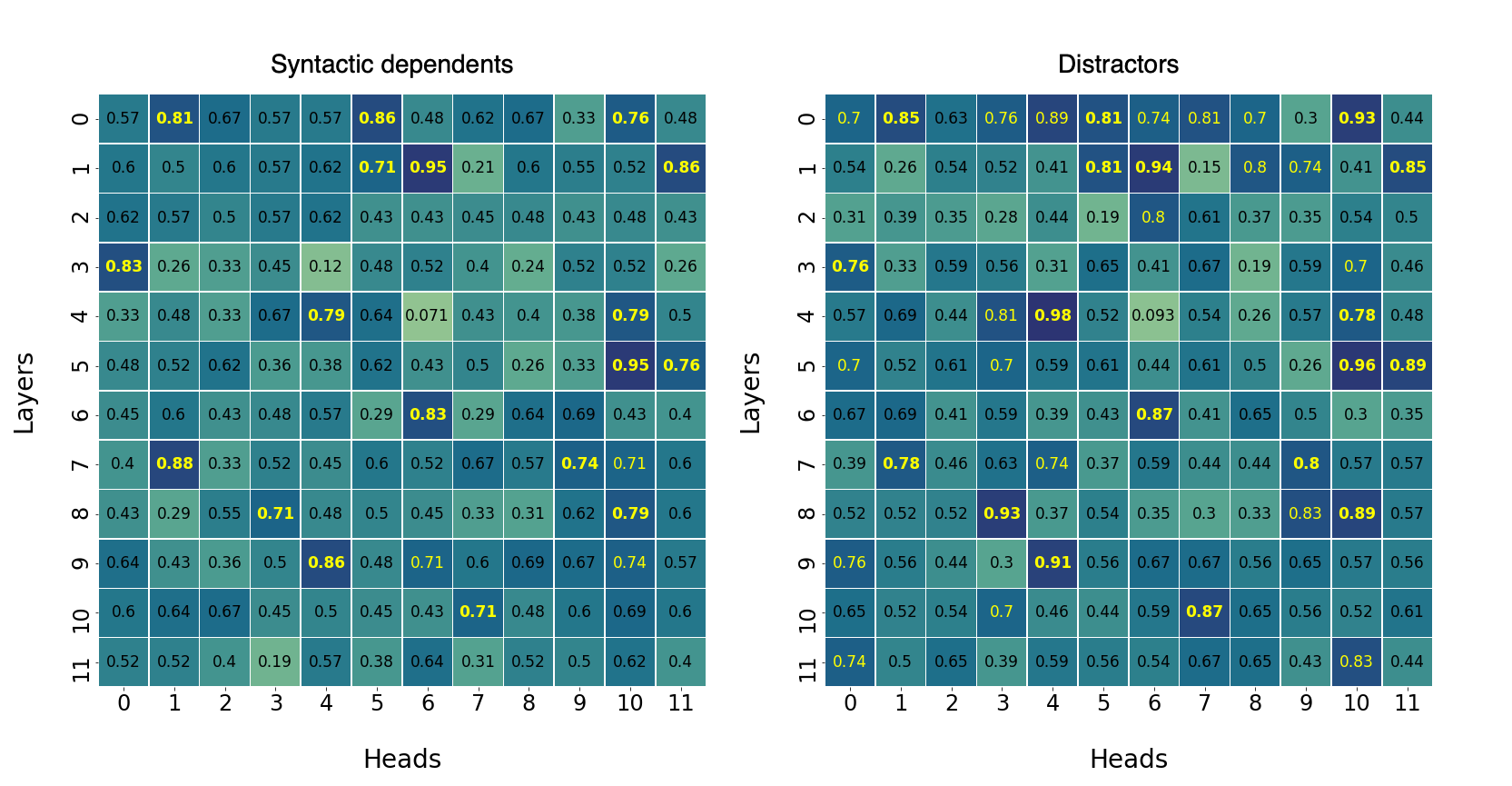}
\begin{center}
    \small{(a) Accuracy}
\end{center}
\vspace{0.1cm}
\includegraphics[scale=0.28]{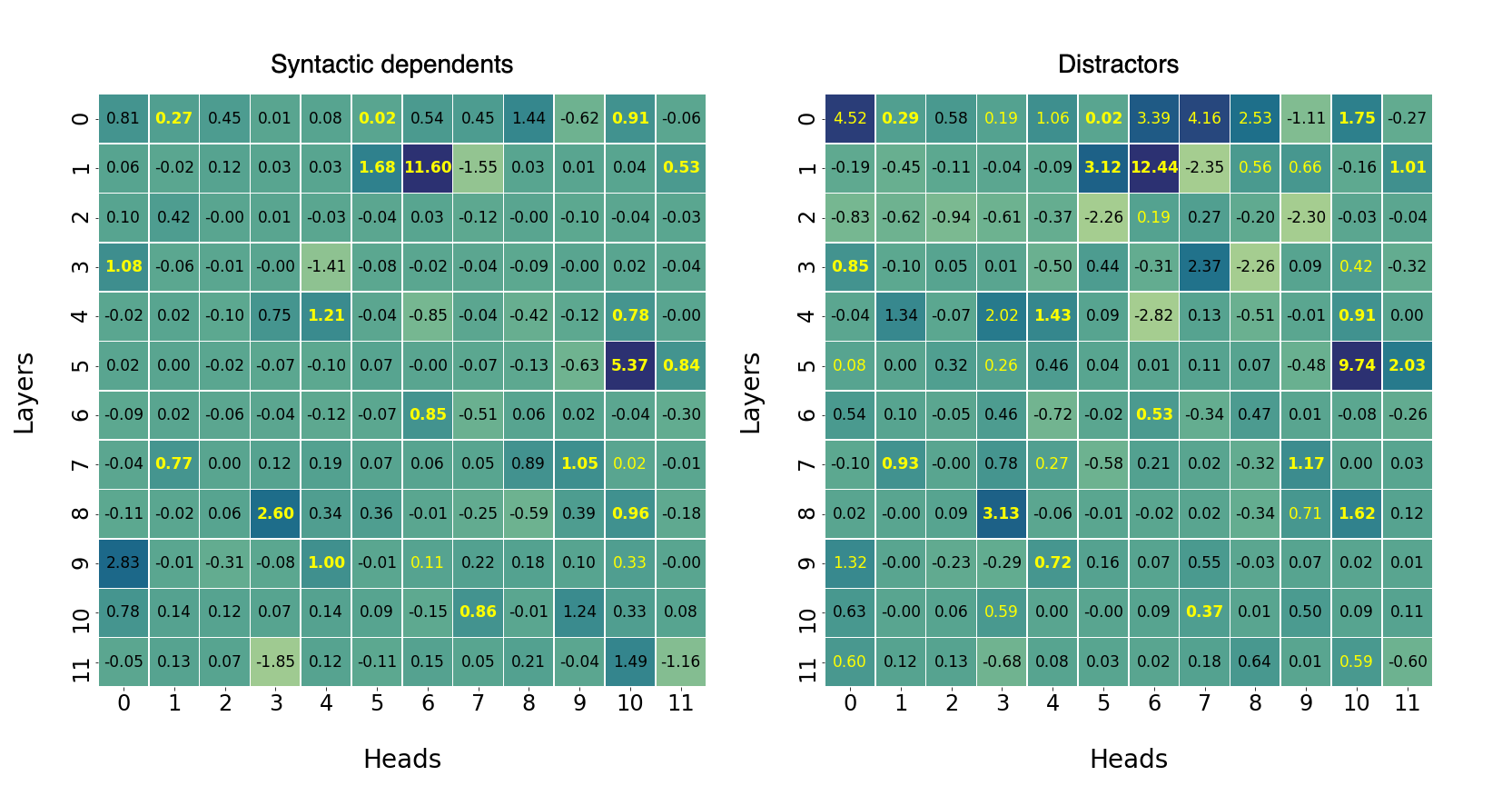}
\begin{center}
    \small{(b) Attention Difference}
\end{center}
\caption{Accuracy and attention difference by attention heads. Attention heads annotated with bold-yellow  are with accuracy greater than 0.70 in both subjects-comparison and distractors-comparison and thus  considered to be specialized for plausibility processing; Attention heads annotated with non-bold-yellow are the ones that showed accuracy greater than 0.70 only for the corresponding condition; Attention heads annotated with black are found to be insensitive to plausibility (accuracies are less than 0.7 for both noun types).}
\label{fig:appendix-a}
\end{figure*}

In addition to accuracy, I also computed attention differences which indicate how much more attention values plausible nouns get compared to implausible nouns (See Equation (\ref{eq:attn-diff})). The attention differences obtained from all attention heads are shown in Figure \ref{fig:appendix-a}.

\begin{equation}\tag{E}
\begin{aligned}
{Attention\ Difference_{lh}}  =   \hspace{2cm}\\
\hspace{-1cm}\sum_{j=1}^{k}[{Attn(pl_{j}, v_{j}) - Attn(impl_{j}, v_{j})]}
\end{aligned}
\label{eq:attn-diff}
\end{equation}
\par
, where \textit{lh} refers to the location of attention heads (\textit{h}th head in the \textit{l}th layer), \textit{j} refers to the sentence id, $pl_{j}$ and $impl_{j}$ refer to the plausible and implausible nouns to be compared in the \textit{j}th sentence set, $v_{j}$ refers to the verb in the \textit{j}th sentence, and \textit{k} is the number of sentence sets.
\par
Metrics were computed two times: one by comparing plausible syntactic dependents and implausible syntactic dependents, and the other by comparing plausible distractors and implausible distractors.

\section{Changes in surprisal values as attention heads are gradually pruned}\label{sec:appendixB}
In Section \ref{subsec:gradual}, it was observed how the plausibility sensitivity changes as the plausibility-processing attention heads are gradually pruned. To provide additional information, this section shows how the surprisals for each condition change along with the gradual head-pruning process.
\par
\begin{figure*}[!t]
\includegraphics[scale=0.35]{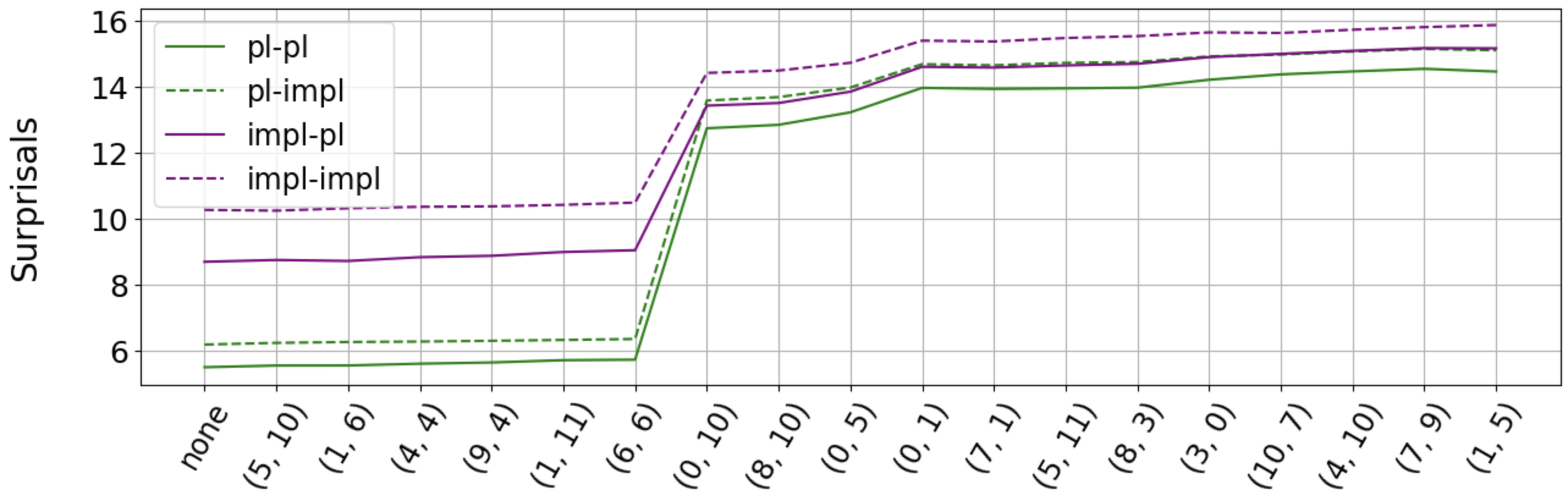}
\caption{Changes in surprisals by conditions as attention heads are gradually pruned. X-axis indicates plausibility processing attention heads that are pruned at a certain point. Attention heads were removed in decreasing order of accuracies in selecting plausible nouns over implausible nouns.}
\label{fig:appendix-b}
\end{figure*}

Surprisals were computed at the verb for each sentence in \citet{cunnings2018retrieval}'s experimental data. The metrics were computed multiple times after removing one of the plausibility-processing attention heads. The computed surprisal values were then averaged by conditions. The plot that shows how surprisal values change by conditions is given in Figure \ref{fig:appendix-b}.

\end{document}